\documentclass{article} 
\usepackage{iclr2026_conference,times}

\usepackage{wrapfig}
\usepackage{lipsum} 
\usepackage{wrapfig}

\usepackage{amsmath,amsfonts,bm}
\usepackage{multicol}

\def\eqref#1{equation~\ref{#1}}

\def\1{\bm{1}}

\DeclareMathAlphabet{\mathsfit}{\encodingdefault}{\sfdefault}{m}{sl}
\SetMathAlphabet{\mathsfit}{bold}{\encodingdefault}{\sfdefault}{bx}{n}

\usepackage{booktabs}      
\usepackage{multirow}      
\usepackage{amssymb}       
\usepackage{graphicx, subfigure}     

\usepackage{colortbl}
\usepackage{xcolor}

\usepackage{hyperref}
\usepackage{url}
\usepackage{multicol}
\usepackage{aliascnt}
\usepackage{adjustbox}
\usepackage{siunitx} 
\usepackage{booktabs}
\usepackage{multirow} 
\usepackage{enumitem}
\usepackage{booktabs} 
\usepackage{amssymb}  
\usepackage{amsmath}  
\usepackage{graphicx} 
\usepackage{longtable} 
\usepackage{graphicx}
\usepackage{subcaption}
\usepackage{calc}
\usepackage[most]{tcolorbox}
\usepackage{listings}

\definecolor{uclablue}{rgb}{0.15, 0.45, 0.68}
\hypersetup{
    breaklinks,
    citecolor=uclablue,
    colorlinks=true,
}

\newtcolorbox{AIbox}[2][]{aibox,title=#2,#1}
\tcbset{
  aibox/.style={
    top=10pt,
    colframe=black,
    colbacktitle=black,
    coltitle=white,
    center,
  }
}

\lstdefinelanguage{prompt}{
    basicstyle=\scriptsize\ttfamily, 
    mathescape=true,        
    escapebegin=\color{latentcolor},  
    escapeend={},
    escapechar=@,
    stringstyle = \color{myorange},
    showstringspaces = false,
    moredelim = [s][\color{mypink}]{`}{`},
    moredelim = [s][\color{mybrown}]{```json}{```},
    moredelim = [s][\color{latentcolor}]{<StartOfLatent>}{<EndOfLatent>},
    literate = %
        {\ \ a.\ }{{\textcolor{mypurple}{\ \ a.\ }}}5
        {\ \ b.\ }{{\textcolor{mypurple}{\ \ b.\ }}}5
        {\ \ c.\ }{{\textcolor{mypurple}{\ \ c.\ }}}5
        {\ \ d.\ }{{\textcolor{mypurple}{\ \ d.\ }}}5
        {\ \ e.\ }{{\textcolor{mypurple}{\ \ e.\ }}}5
        {\ \ f.\ }{{\textcolor{mypurple}{\ \ f.\ }}}5
        {\ \ g.\ }{{\textcolor{mypurple}{\ \ g.\ }}}5
        {\ \ h.\ }{{\textcolor{mypurple}{\ \ h.\ }}}5
        {\ I.\ }{{\textcolor{mypurple}{\ I.\ }}}4
        {\ II.\ }{{\textcolor{mypurple}{\ II.\ }}}5
        {\ III.\ }{{\textcolor{mypurple}{\ III.\ }}}6
        {\ IV.\ }{{\textcolor{mypurple}{\ IV.\ }}}5
        {\ V.\ }{{\textcolor{mypurple}{\ V.\ }}}4
}

\newtcblisting[auto counter, number within=subsection]{prompt}[4][]{%
  before skip=6pt, after skip=6pt,    
  top=6pt, bottom=6pt, left=7pt, right=7pt, 
  enhanced, breakable,
  colback=white, colframe=gray!65,
  boxrule=0.6pt, arc=2pt,
  drop shadow={black!15},             
  listing only,
  listing options={
    language=prompt,
    upquote=true,
    basicstyle=\scriptsize\ttfamily \setlength{\baselineskip}{1.1\baselineskip},
    columns=fullflexible,
    breaklines=true,
    breakindent=0pt,
    xleftmargin=\ifx\empty#2\empty0pt\else#2\fi,
    xrightmargin=\ifx\empty#3\empty0pt\else#3\fi,
    aboveskip=0pt,                    
    belowskip=0pt,                    
    showstringspaces=false,
  },
  title={\ifstrempty{#4}{Prompt \thetcbcounter}{Prompt \thetcbcounter: #4}},
  colbacktitle=gray!6, coltitle=black,
  attach boxed title to top left={yshift=-1mm, xshift=6pt},
  boxed title style={colback=gray!6, boxrule=0pt, sharp corners},
  #1
}

\newtcblisting[auto counter, number within=subsection]{showcase}[4][]{%
  before=\par\vspace{\baselineskip},
  after=\par,
  width=\linewidth,
  enhanced,
  arc=0em,
  boxrule=1pt,
  listing only,
  listing options={
    language=prompt,
    upquote=true,
    basicstyle=\scriptsize\ttfamily \setlength{\baselineskip}{1.1\baselineskip},
    breaklines=true,
    breakindent=0pt,
    xleftmargin=\ifx\empty#2\empty-12pt\else#2\fi,
    xrightmargin=\ifx\empty#3\empty-5pt\else#3\fi,
    aboveskip=-4pt,
    belowskip=-4pt,
    columns=fullflexible,
    },
  colback=white,
  colframe=gray,
  colbacktitle=gray!5,
  coltitle=black,
  attach boxed title to top center={yshift=-3mm},
  box align=center,
  parbox=false,
  title={\ifstrempty{#4}{Example \thetcbcounter}{Example \thetcbcounter: #4}},
  #1
}

\definecolor{linkColor}{rgb}{0.2,0.4,0.6}
\definecolor{myblue}{HTML}{0379AC}
\definecolor{myred}{HTML}{A50E50}
\definecolor{myorange}{RGB}{238, 133, 74}
\definecolor{latentcolor}{named}{cyan}
\definecolor{normalcolor}{RGB}{0, 0, 0}
\usepackage{marvosym}

\newcommand{\ours}{\textbf{WRAP++}}

\title{WRAP++: Web Discovery Amplified Pretraining}

\author{
Jiang Zhou, Yunhao Wang, Xing Wu$^\dagger$, Tinghao Yu, Feng Zhang
}

\iclrfinalcopy
\begin{document}
\maketitle
\let\oldthefootnote\thefootnote

\let\thefootnote\relax\footnotetext{$^\dagger$ Corresponding Author. Correspondence to ucaswu@tencent.com.}

\let\thefootnote\relax\footnotetext{}
\let\thefootnote\oldthefootnote

\begin{abstract}
Synthetic data rephrasing has emerged as a powerful technique for enhancing knowledge acquisition during large language model (LLM) pretraining.
However, existing approaches operate at the \emph{single-document} level, rewriting individual web pages in isolation.
This confines synthesized examples to intra-document knowledge, missing cross-document relationships and leaving facts with limited associative context.
We propose \ours{} (\textbf{W}eb discove\textbf{R}y \textbf{A}mplified \textbf{P}retraining), which \emph{amplifies} the associative context of factual knowledge by \emph{discovering} cross-document relationships from web hyperlinks and synthesizing joint QA over each discovered document pair.
Concretely, \ours{} discovers high-confidence relational motifs including dual-links ($A \leftrightarrow B$) and co-mentions ($A \rightarrow E \leftarrow B$ with $A \rightarrow B$), and synthesizes QA that requires reasoning across both documents. This produces relational knowledge absent from either source document alone, creating diverse entry points to the same facts.
Because the number of valid entity pairs grows combinatorially, this discovery-driven synthesis also amplifies data scale far beyond single-document rewriting.
Instantiating \ours{} on Wikipedia, we amplify $\sim$8.4B tokens of raw text into 80B tokens of cross-document QA data. On SimpleQA, OLMo-based models at both 7B and 32B scales trained with \ours{} substantially outperform single-document approaches and exhibit sustained scaling gains, underscoring the advantage of cross-document knowledge discovery and amplification.

\end{abstract}

\section{Introduction}
\label{sec:introduction}

Synthetic data has become an increasingly important component of large language model (LLM) pretraining. WRAP~\citep{maini2024wrap} showed that rephrasing noisy web text into QA format can improve pretraining, and later systems scaled this recipe substantially: Nemotron-CC~\citep{su-etal-2025-nemotron} produced $\sim$2 trillion synthetic tokens from Common Crawl, Phi-4~\citep{abdin2024phi4technicalreport} used 40\% synthetic data in pretraining, and Qwen3~\citep{yang2025qwen3} incorporated synthetic data into its training pipeline.

However, this progress has been explored mainly along intra-document axes---rephrasing strategy, generator model, and source quality~\citep{nguyen2025recycling,niklaus2026finephrase}---varying \emph{how} a single document is rewritten without changing \emph{what} is synthesized.
Because many facts are distributed across multiple documents, this single-document paradigm confines the model to limited associative context for each fact, which ultimately hinders knowledge recoverability.

This limitation motivates a shift toward \emph{cross-document synthesis}: bringing multiple facts into a shared context to learn relational knowledge jointly.
However, moving to cross-document synthesis is non-trivial.
A na\"{i}ve approach of randomly pairing documents yields little improvement over single-document baselines (see \S\ref{sec:ablation_graph}), as forcing an LLM to synthesize joint QA from unrelated texts produces fabricated connections and low-quality data.
Thus, the \emph{document selection} mechanism is critical: cross-document synthesis only succeeds when the paired documents contain genuinely related facts.

\begin{figure}
    \centering
    \includegraphics[width=1\linewidth]{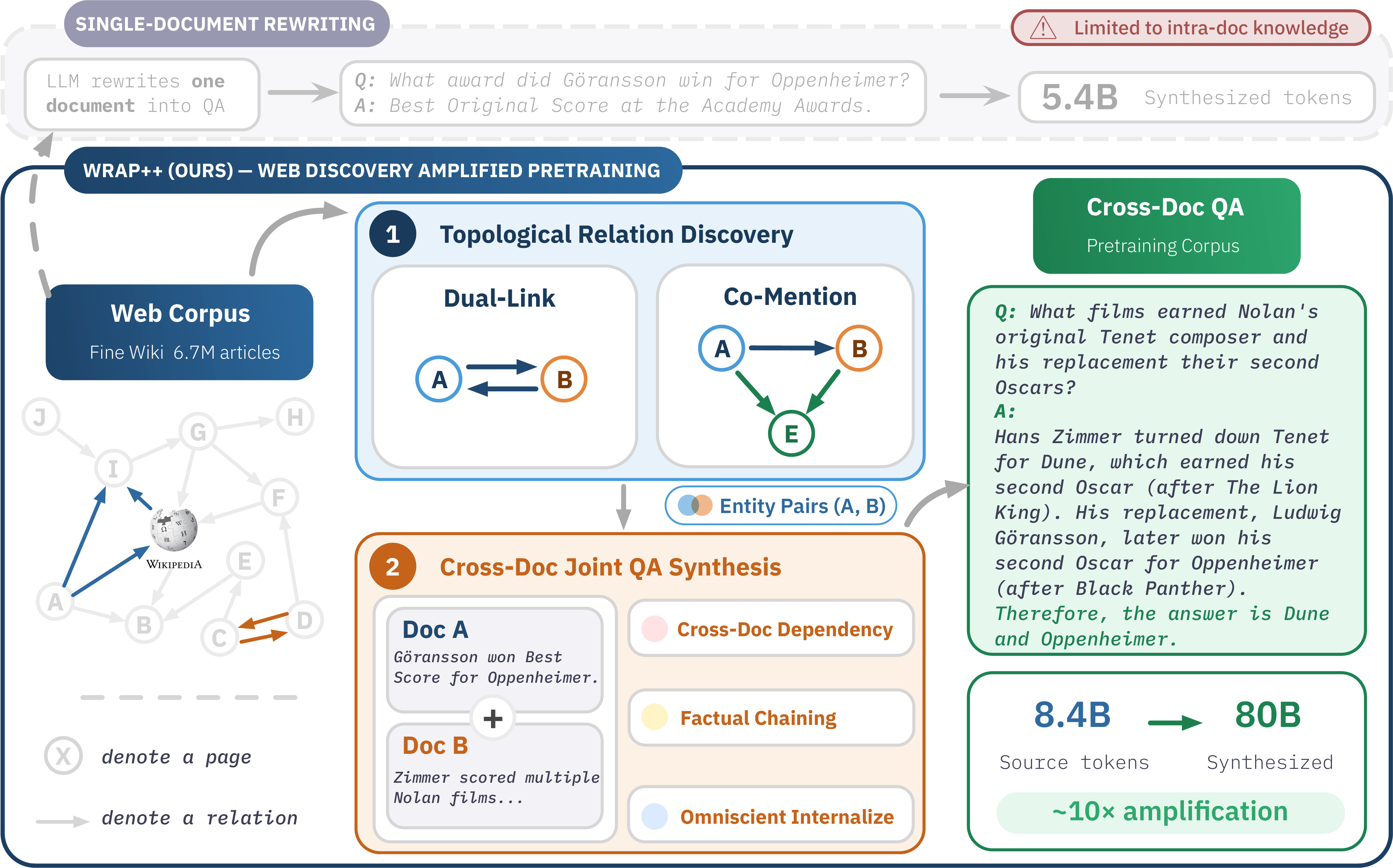}
    \caption{Overview of the \ours{} pipeline. Unlike single-document WRAP, which rewrites individual documents, \ours{} \emph{discovers} cross-document entity relationships from web topology and \emph{amplifies} them into pretraining data through joint QA synthesis.}
\label{fig:method_overview}
\end{figure}

Web hyperlinks provide a broad relevance signal by encoding human-curated judgments of importance~\citep{zhou2022hyperlink}. 
For example, the Wikipedia pages of composers Hans Zimmer and Ludwig G\"{o}ransson are topologically linked through shared collaborations with director Christopher Nolan. While single-document synthesis might only extract isolated facts (e.g., "G\"{o}ransson won an Oscar for Oppenheimer"), \ours{} pairs these connected documents to synthesize multi-hop relational QA. As illustrated in Figure \ref{fig:method_overview}, the model is forced to explicitly reason across both texts—deducing that Zimmer left Tenet to score Dune (his second Oscar), leading to G\"{o}ransson's hiring and subsequent second Oscar for Oppenheimer. This explicit relational supervision provides vital disambiguation cues and diverse retrieval paths absent from single-document synthesis, saving the LLM from having to implicitly infer these complex connections from massive unstructured text.

\label{sec:method}

Based on this insight, we propose \ours{} (\textbf{W}eb discov\textbf{R}y \textbf{A}mplified \textbf{P}retraining), which effectively extends the synthetic data paradigm from single-document \emph{rewriting} to cross-document \emph{discovery and amplification}. 

\textbf{Amplifying Associative Context via Relation Discovery.} We discover high-confidence relational motifs from web hyperlinks~\citep{zhou2022hyperlink}---specifically \emph{dual-links} ($A \leftrightarrow B$) and \emph{co-mentions} ($A \rightarrow E \leftarrow B$ with $A \rightarrow B$). Rather than simply concatenating these documents, we feed the discovered pairs to an instruction-tuned LLM generator subjected to three strict synthesis constraints: enforcing \textbf{Cross-Document Dependency} to mandate joint reasoning, requiring \textbf{Explicit Factual Chaining} to decode multi-hop logical paths, and ensuring \textbf{Omniscient Internalization} by forbidding local document attribution. This process produces genuinely new relational knowledge (comparisons, contrasts, bridging facts) that creates diverse retrieval paths to the same facts. Furthermore, because the number of valid entity pairs grows combinatorially, this discovery-driven synthesis achieves a $\sim$10$\times$ data amplification---scaling a fixed 8.4B-token source corpus into 80B tokens of cross-document QA data---consistently improving the knowledge recoverability of the model.

We instantiate \ours{} on Wikipedia and amplify $\sim$8.4B tokens of raw text into 80B tokens of cross-document QA data---compared to only $\sim$5.4B tokens from single-document WRAP.
On the SimpleQA benchmark~\citep{wei2024simpleqa}, OLMo-based models at both 7B and 32B scales trained with \ours{} data substantially outperform all single-document baselines, and \ours{} demonstrates a more favorable scaling trajectory than single-document approaches.

Our contributions are three-fold:

\begin{enumerate}
    \item We propose \ours{}, a framework that extends single-document rewriting into topology-guided relation discovery and joint QA synthesis, amplifying the associative context of factual knowledge.
    \item We instantiate \ours{} on Wikipedia to synthesize 80B tokens of cross-document QA data, demonstrating that combinatorial relation discovery enables data amplification far beyond single-document synthesis.
    \item We show on SimpleQA with OLMo-based 7B and 32B models that \ours{} substantially outperforms single-document baselines, exhibiting a favorable scaling trajectory.
\end{enumerate}

\section{Method: \ours{}}

\ours{} is a framework that transitions synthetic pretraining data from single-document rewriting to topology-guided cross-document discovery and amplification. The framework consists of two core stages: Topological Relation Discovery (\S\ref{sec:discovery}) and Joint QA Synthesis (\S\ref{sec:synthesis}). Figure~\ref{fig:method_overview} illustrates the overall pipeline.

\subsection{Problem Formulation and Graph Abstraction}
\label{sec:formulation}

Let $\mathcal{D} = \{d_1, d_2, \dots, d_N\}$ denote a large-scale web corpus consisting of $N$ documents. The inter-document references (e.g., hyperlinks) naturally induce a directed knowledge graph $\mathcal{G} = (\mathcal{V}, \mathcal{E})$, where each vertex $v_i \in \mathcal{V}$ corresponds to a document $d_i$, and a directed edge $e_{i,j} \in \mathcal{E}$ exists if $d_i$ explicitly references $d_j$. Since each document in our corpus describes a single entity, we use ``entity pair'' and ``document pair'' interchangeably throughout this paper.

Conventional single-document synthesis operates solely on the local context of $v_i$, limiting the model's exposure to isolated facts. In contrast, \ours{} leverages the topological structure of $\mathcal{G}$ to discover genuine semantic dependencies across documents, bringing related knowledge into a shared synthesis context to amplify the associative context of factual knowledge.

\subsection{Topological Relation Discovery}
\label{sec:discovery}

A na\"{i}ve approach of pairing random documents from $\mathcal{D}$ forces the synthesis model to hallucinate spurious connections. To ensure the semantic validity of cross-document synthesis, we discover high-confidence relational motifs directly from $\mathcal{G}$. We focus on two topological structures that provide strong inductive biases for relational reasoning:

\paragraph{Dual-link Motif.}
Two documents $u$ and $v$ form a dual-link relationship if they mutually reference each other. Formally, a dual-link pair $(u, v)$ satisfies the bidirectional constraint:
\begin{equation}
    e_{u,v} \in \mathcal{E} \land e_{v,u} \in \mathcal{E}
\end{equation}
This mutual dependency typically indicates a strong, foundational semantic correlation (e.g., a notable director and their magnum opus, or a scientist and their core discovery). Discovering this motif ensures the underlying entity pair is highly coupled.

\paragraph{Co-mention Motif.}
Documents $u$ and $v$ share a co-mention relationship if they both reference a common structural hub $E$, while maintaining a direct link between themselves. Formally, the triplet $(u, v, E)$ satisfies:
\begin{equation}
    e_{u,E} \in \mathcal{E} \land e_{v,E} \in \mathcal{E} \land e_{u,v} \in \mathcal{E}
\end{equation}
The shared structural context $E$ imposes implicit analogical, hierarchical, or comparative relationships (e.g., two competing theories cited in the same survey article). This motif explicitly encourages the subsequent synthesis model to generate relational knowledge that contrasts and compares the related entities, thereby amplifying their associative context.

\subsection{Cross-Document Joint QA Synthesis}
\label{sec:synthesis}

Given a discovered document pair $(d_u, d_v)$ connected by a valid topological motif, we employ an instruction-tuned LLM generator $\mathcal{M}_{\theta}$ to synthesize a set of composite QA instances $\mathcal{S}_{u,v} = \{(q_i, c_i, a_i)\}_{i=1}^K$, where $q_i$ is the question, $c_i$ is the intermediate factual chain, and $a_i$ is the final answer.

The generation process is conditioned on a structured prompt $\mathcal{P}$ and the concatenated document context:
\begin{equation}
    \mathcal{S}_{u,v} \sim \mathcal{M}_{\theta}(d_u \oplus d_v, \mathcal{P})
\end{equation}

To amplify the associative context of the synthesized data and prevent the generator from degrading to shallow single-document summarization, $\mathcal{P}$ enforces three functional constraints on the output space:

\begin{itemize}
    \item \textbf{Strict Cross-Document Dependency:} The generated question $q_i$ must have high entropy given only one document. Deriving the correct answer $a_i$ must strictly require logical premises from both $d_u$ and $d_v$, ensuring the synthesis produces genuinely new relational knowledge rather than merely rephrasing isolated facts.
    \item \textbf{Explicit Factual Chaining:} Before outputting $a_i$, the generator must explicitly decode the traversal path $c_i$. By articulating the necessary facts extracted from both documents and linking them step-by-step, the pretraining model internalizes multi-hop knowledge structures, effectively creating diverse associative entry points to the underlying facts.
    \item \textbf{Omniscient Internalization:} The generator is strictly prohibited from attributing facts to the local context (e.g., avoiding ``According to Passage A''). It must output universally valid statements. This ensures the synthesized data serves as parametric world knowledge rather than context-dependent reading comprehension exercises.
\end{itemize}

\section{Experiments}
\label{sec:experiments}

\subsection{Experimental Setup}
\label{sec:instantiation}

\paragraph{Synthesis Data.}
In principle, \ours{} is a general framework applicable to any text corpus containing hyperlinks. In this work, we instantiate it on Wikipedia because of its rich link structure and its widespread use in prior work on synthetic rewriting methods \citep{maini2024wrap,su-etal-2025-nemotron}. Specifically, we use the English subset of FineWiki~\citep{penedo2025finewiki} as our base corpus ($\mathcal{D}$), which contains approximately 8.4B tokens. We parse the hyperlinks in FineWiki to construct the directed inter-document graph $\mathcal{G}$ used for topological relation discovery.

\paragraph{Synthesis Model.}
We use Qwen3-30B-A3B-Instruct-FP8 as our instruction-tuned generator $\mathcal{M}_{\theta}$. The prompt is designed to enforce the strict cross-document dependency and explicit factual chaining constraints described in \S\ref{sec:method}, thereby encouraging high-quality relational QA generation. The full prompt template is provided in Appendix~\ref{app:prompts}.

\paragraph{Data Scale.}
Topological relation discovery substantially amplifies the data scale beyond individual documents. The dual-link motif yields highly coupled entity pairs that produce $\sim$3B tokens of cross-document QA data. Incorporating the co-mention motif broadens coverage, bringing the combined \ours{} dataset to $\sim$82.7B tokens. 

\paragraph{Training Models.}
To assess the effect of cross-document synthetic data on parametric knowledge, we continue pretraining from the OLMo-3 stage-1 last checkpoint at both the 7B and 32B scales for one epoch. We choose OLMo because it is fully open-source and provides publicly released checkpoints throughout training, making it a suitable platform for controlled continued-pretraining experiments.

\subsection{Evaluation Setup}
\label{sec:eval_setup}

\paragraph{Benchmark.}
We evaluate on SimpleQA~\citep{wei2024simpleqa}, a knowledge-intensive benchmark designed to measure short-form factual accuracy while minimizing sensitivity to formatting heuristics. Most SimpleQA questions can be answered directly from Wikipedia-derived knowledge, making it a natural testbed for studying factual knowledge acquisition under our setup.

\paragraph{Metric.}
We use pass@128 as our primary metric, defined as the empirical probability that at least one of 128 sampled responses contains the correct fact. Our goal is to measure \emph{knowledge recoverability} rather than only top-1 answer accuracy. In this setting, pass@128 is useful because it probes whether the correct factual association can be elicited from the model under sampling, even when it is not the single most likely surface form. We therefore treat it as a more sensitive indicator of parametric knowledge recoverability during continued pretraining than pass@1.

\paragraph{Baselines.}
To isolate the benefit of cross-document synthesis, we compare \ours{} against two closely related single-document baselines derived from the same FineWiki corpus:
\begin{enumerate}
    \item \textbf{WRAP ($\sim$5.4B tokens)}: Standard single-document QA synthesis following the original WRAP recipe~\citep{maini2024wrap}, representing the typical yield of intra-document fact extraction.
    \item \textbf{Extended WRAP ($\sim$17.4B tokens)}: An expanded single-document synthesis utilizing additional prompting strategies (e.g., exhaustive extraction) to push the limits of single-document scaling.
\end{enumerate}
These baselines also illustrate the data-scaling constraint of single-document synthesis: on the same FineWiki corpus, standard WRAP yields only $\sim$5.4B tokens and Extended WRAP reaches $\sim$17.4B tokens, both well below the $\sim$82.7B-token scale of \ours{}.

\subsection{Main Results}

\begin{table}[t]
\centering
\caption{SimpleQA results after 1-epoch continued training on OLMo-3-7B and OLMo-3-32B using different data recipes. The metric reported is the empirical pass@128 rate (\%).}
\label{tab:main_results}
\begin{tabular}{lcc}
\toprule
\textbf{Data Recipe} & \textbf{OLMo-3-7B} & \textbf{OLMo-3-32B} \\
\midrule
\textit{Pretrained Base} & 34.76 & 42.35 \\
\midrule
+ WRAP                   & 39.55 & 44.43 \\
+ Extended WRAP          & 43.69 & 47.91 \\
\rowcolor{gray!10}
+ \ours{}                & \textbf{49.13} & \textbf{53.97} \\
\bottomrule
\end{tabular}
\end{table}

Table~\ref{tab:main_results} presents the main results of 1-epoch continued training across different synthesis recipes. We highlight two principal findings.

\paragraph{\ours{} substantially outperforms single-document baselines.}
Across both model scales, continued pretraining with \ours{} yields substantially higher pass@128 on SimpleQA compared to all single-document approaches (\textit{+9.5 pp on 7B, +9.8 pp on 32B over WRAP; +5.4 pp on 7B, +6.1 pp on 32B over Extended WRAP}).
This advantage reflects two complementary factors.
First, cross-document synthesis produces higher-quality relational knowledge \emph{per token}: at a matched budget of $\sim$8B tokens, \ours{} already outperforms Extended WRAP by +2.48 pp (detailed in \S\ref{sec:ablation_strategies}), confirming a genuine quality advantage independent of data scale.
Second, the combinatorial nature of relation discovery amplifies this quality advantage to a far larger data space ($\sim$80B tokens) that single-document methods cannot access, yielding further gains as training progresses (Figure~\ref{fig:scaling_curve}).

\paragraph{Surpassing the single-document scaling bottleneck.}
Single-document methods face an inherent data bottleneck: the finite number of extractable facts within an individual page. While Extended WRAP attempts to push this limit through exhaustive extraction (reaching $\sim$17.4B tokens), it ultimately depletes the source material. The resulting diminishing returns (+4.1 pp on 7B, +3.7 pp on 32B over standard WRAP) suggest information saturation under the single-document paradigm.
In contrast, because the number of valid cross-document entity pairs grows combinatorially, \ours{} amplifies the same FineWiki source corpus into $\sim$80B tokens of relational knowledge---a data space fundamentally inaccessible to single-document methods. Single-document methods \emph{cannot} close this gap simply by training longer, since their source material is already exhausted. We analyze the resulting scaling dynamics in detail next.

\subsection{Scaling and Training Dynamics}
\label{sec:scaling_dynamics}

\begin{figure}[t]
    \centering
    \includegraphics[width=0.85\linewidth]{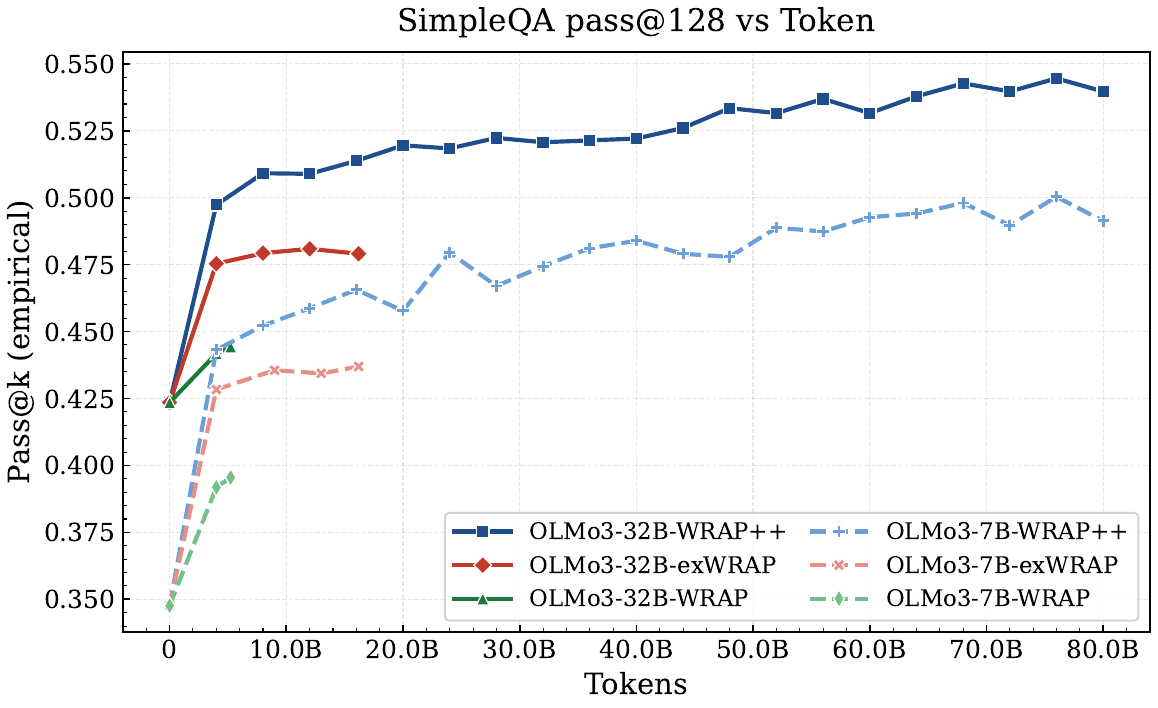}
    \caption{\textbf{SimpleQA pass@128 vs.\ training tokens.} Single-document recipes (WRAP and Extended WRAP) reach a data bottleneck early, limiting further knowledge acquisition. In contrast, the combinatorial nature of \ours{} allows it to scale effectively up to 80B tokens, improving performance without obviously plateauing.}
    \label{fig:scaling_curve}
\end{figure}

\begin{figure}[t]
    \centering
    \includegraphics[width=0.85\linewidth]{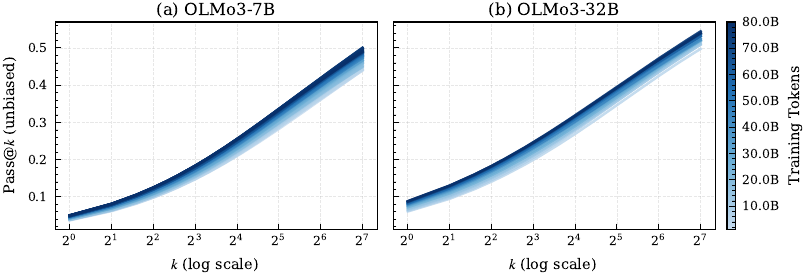}
    \caption{\textbf{Evolution of pass@$k$ performance during training.} The curves illustrate the unbiased pass@$k$ of OLMo-3-7B (a) and OLMo-3-32B (b) on SimpleQA ($k \in [1, 128]$ in log scale). The color gradient (from light to dark blue) tracks the accumulation of consumed \ours{} tokens (from 10B to 80B). The strictly monotonic upward shift across all values of $k$ indicates robust, unsaturated knowledge internalization.}
    \label{fig:training_dynamics}
\end{figure}

Figure~\ref{fig:scaling_curve} plots pass@128 as a function of training tokens consumed. The trajectories confirm the scaling bottleneck discussed above: single-document recipes plateau early, whereas \ours{} maintains a steady upward trend all the way to 80B tokens without obvious saturation, demonstrating that the combinatorial data space opened by relation discovery translates into sustained knowledge gains.

To further dissect how this scaling translates into knowledge recoverability, we track the evolution of the pass@$k$ curves throughout training. Figure~\ref{fig:training_dynamics} visualizes the unbiased SimpleQA pass@$k$ for both models. As training progresses (indicated by the light-to-dark blue gradient representing the 80B token influx), the curves exhibit a monotonic upward shift across all values of $k$.

Crucially, this improvement spans the entire logarithmic $k$-spectrum. The persistent lift at small $k$ (the leftmost regions of the curves) shows that the model's top-ranked answers increasingly contain the correct fact, reflecting higher precision. Simultaneously, the parallel gains at larger $k$ indicate a broader and more robust set of associative retrieval paths to the same knowledge.

\section{Ablations and Analysis}
\label{sec:ablations}

We conduct extensive ablations to validate each component of \ours{}.
Due to experimental costs, unless otherwise noted, all ablation experiments use OLMo-3-7B continued pretraining with $\sim$8B tokens and report the results in terms of SimpleQA pass@128.

\begin{table}[t]
\centering
\caption{Ablation results across different design choices (OLMo-3-7B, $\sim$8B tokens). We report SimpleQA pass@128. \ours{} (default) uses topological relation discovery, combined topologies, joint QA synthesis, and the Qwen3-30B-A3B synthesis model.}
\label{tab:ablation_combined}
\scalebox{0.85}{
\begin{tabular}{llc}
\toprule
\textbf{Ablation Axis} & \textbf{Variant} & \textbf{pass@128} \\
\midrule
\multirow{2}{*}{Pairing Strategy (\S\ref{sec:ablation_graph})} & Random entity pairing & 43.46 \\
& \textbf{Topological relation discovery} & \textbf{45.11} \\
\midrule
\multirow{3}{*}{Relation Topology (\S\ref{sec:ablation_topology})} & Dual-link only & 44.24 \\
& Co-mention only & 44.36 \\
& \textbf{Combined} & \textbf{45.11} \\
\midrule
\multirow{3}{*}{Synthesis Method (\S\ref{sec:ablation_synthesis})} & Raw concatenation (no QA) & 35.43 \\
& QA with source documents prepended & 38.93 \\
& \textbf{Joint QA synthesis} & \textbf{45.11} \\
\midrule
\multirow{2}{*}{Synthesis Model Scale (\S\ref{sec:ablation_model})} & \textbf{Qwen3-30B-A3B} & 45.11 \\
& Qwen3-235B-A22B & \textbf{47.70} \\
\bottomrule
\end{tabular}}
\end{table}

\subsection{Necessity of Topological Relation Discovery}
\label{sec:ablation_graph}

We explore whether the topological relation discovery is essential, or whether randomly pairing Wikipedia pages would suffice.
Table~\ref{tab:ablation_combined} (Pairing Strategy rows) shows a clear performance drop (from 45.11 to 43.46) when entities are paired randomly rather than via dual-link or co-mention relations.
Qualitatively, random pairing forces the synthesis model to fabricate relationships between unrelated entities, producing factually incorrect comparisons and superficial connections.
This confirms that principled relation discovery---specifically, dual-link and co-mention motif discovery---is important for high-quality cross-document synthesis.

\subsection{Topology Comparison: Dual-Link vs.\ Co-Mention}
\label{sec:ablation_topology}

We explore the contribution of each relation type at a matched token budget of $\sim$8B.
As shown in Table~\ref{tab:ablation_combined} (Relation Topology rows), both topologies provide strong relational signal at this budget. Co-mention retains a slight edge over dual-link (44.36 vs.\ 44.24), while their combination yields the best overall performance (45.11). This suggests that bidirectional links and shared structural context capture yet complementary aspects of cross-document knowledge.

\subsection{Necessity and Format of QA Synthesis}
\label{sec:ablation_synthesis}

We explore the optimal data format for learning cross-document relationships by comparing our joint QA synthesis against two alternatives: (1) raw concatenation of related documents (no QA), and (2) prepending source documents to the synthesized QA pairs. Table~\ref{tab:ablation_combined} shows raw concatenation performs only marginally above the pretrained base (35.43 vs.\ 34.8), indicating that explicit \emph{synthesis} is essential to convert document proximity into learnable relational knowledge. Moreover, prepending source documents to QA pairs causes a notable performance drop (38.93), likely by allowing the model to superficially copy answers rather than parametrically internalizing them. Thus, joint QA synthesis provides the most effective format for amplifying associative context.

\subsection{Effect of Synthesis Model Scale}
\label{sec:ablation_model}

We explore the effect of synthesis model scale on \ours{} quality by comparing two synthesis models of different scales: Qwen3-30B-A3B (3B active parameters) and Qwen3-235B-A22B (22B active parameters).
Table~\ref{tab:ablation_combined} (Synthesis Model Scale rows) shows that the larger model produces higher-quality cross-document QA, leading to better downstream pass@128.
In practice, the choice between synthesis models involves a cost--quality tradeoff: the larger model is preferable when generation budget is not the bottleneck, while the smaller model enables broader coverage at lower compute cost.

\begin{table}[t]
\centering
\caption{Performance comparison of \ours{} mixed with other single-document strategies. Each mixture contains $\sim$8B tokens total.}
\label{tab:ablation_mix_128}
\scalebox{0.85}{
\begin{tabular}{lcc}
\toprule
\multirow{2}{*}{Single-Document Component} & \multicolumn{2}{c}{Proportion of \ours{} in Mixture} \\
\cmidrule(lr){2-3}
& 0\% (Baseline Only) & 50\% (1:1 Mix) \\
\midrule
\textit{Pretrained Model (No further training)} & \multicolumn{2}{c}{34.76}  \\
\midrule
Raw FineWiki       & 39.23 & 41.80 \\
Distill         & 38.12 & 41.59 \\
Extract Knowledge     & 38.69 & 43.35 \\
Knowledge List   & 38.60 & 42.03 \\
\midrule
\textbf{\ours{} (100\%, Ours)}    & \multicolumn{2}{c}{45.11} \\
\bottomrule
\end{tabular}}
\end{table}

\subsection{Comparison with Other Single-Document Strategies}
\label{sec:ablation_strategies}

We contextualize the performance of \ours{} against other representative single-document rephrasing strategies \citet{su-etal-2025-nemotron} applied to the identical FineWiki source corpus, including:
(a) \textbf{Distill}---rewriting into cleaner, more concise prose while preserving information;
(b) \textbf{Extract Knowledge}---extracting key factual statements and discarding redundancy;
(c) \textbf{Knowledge List}---outputting structured knowledge in list format.

Table~\ref{tab:ablation_mix_128} presents both the isolated performance (0\% and 100\%) and the mixing dynamics (50\% blending) at a strictly restricted budget of $\sim$8B tokens. 
When evaluated in isolation at the 8B-token budget, pure \ours{} (45.11) substantially outperforms all listed single-document baselines. This margin at a restricted data scale reveals an important dynamic in pretraining efficiency: discovery-driven synthesis already yields stronger knowledge recoverability, even before exploiting its data amplification headroom.
When blending \ours{} with other strategies in a 1:1 ratio, we observe a clear \emph{uplift effect}: injecting \ours{} into any weaker baseline consistently improves upon its 0\% counterpart (e.g., Knowledge List rises from 38.60 to 42.03), further highlighting the advantage of discovery-driven synthesis.

\subsection{Integration with OLMo-3 Mid-Training Data}
\label{sec:ablation_midtrain}

\begin{table}[t]
    \centering
    \caption{Integrating \ours{} into OLMo-3-7B 100B-token mid-training. We report SimpleQA pass@128 and the average over 12 general benchmarks (including MMLU Redux, HellaSwag, etc. See Appendix~\ref{app:midtrain_breakdown} for the full list of tasks).}
    \label{tab:ablation_midtrain}
    \begin{tabular}{lcc}
    \toprule
    \textbf{Setting} & \textbf{SimpleQA} & \textbf{Gen. Avg} \\
    \midrule
    & pass@128 & (12 tasks) \\
    \midrule
    \textit{Pretrained Base} & 34.76 & 57.79 \\
    + Midtrain (100B) & 34.74 & \textbf{68.24} \\
    + \ours{} Mix (100B) & \textbf{37.58} & 68.16 \\
    \bottomrule
    \end{tabular}
\end{table}
    
We further explore whether \ours{} data can be integrated into a realistic mid-training pipeline without harming general capabilities.
We augment OLMo-3's 100B-token mid-training mixture with 6B tokens of \ours{} data and train for the full schedule. As a baseline, we train on the original OLMo-3 mid-training mixture under identical conditions.
Following \citet{niklaus2026finephrase}, we evaluate both SimpleQA and the average performance across 12 general tasks (detailed in Appendix~\ref{app:midtrain_breakdown}).
Table~\ref{tab:ablation_midtrain} shows that adding \ours{} data yields a meaningful improvement on SimpleQA (+2.9 points pass@128) while maintaining a comparable general-benchmark average (68.16 vs.\ 68.24).
Notably, \ours{} explicitly enhances knowledge-intensive tasks, yielding clear gains on MMLU Redux (+1.28, detailed in Appendix~\ref{app:midtrain_breakdown}).
This demonstrates that \ours{} integrates cleanly into full-scale mid-training, preserving broad capabilities while explicitly strengthening the model's general knowledge foundation. Moreover, as the mid-training budgets of leading models grow toward the trillion-token regime \citep{yang2025qwen3,zeng2026glm}, the 80B-token scale of \ours{} suggests strong potential for integration into future large-scale training pipelines.

\section{Related Work}
\label{sec:related}

\paragraph{Synthetic Data for LLM Pretraining.}
WRAP~\citep{maini2024wrap} established synthetic rephrasing as a practical pretraining paradigm, showing that rewriting web documents into cleaner QA-style text with instruction-tuned models can accelerate pretraining by $\sim$3$\times$. Subsequent work has expanded this design space along three main axes. First, on \emph{rephrasing strategy}, Nemotron-CC~\citep{su-etal-2025-nemotron} extracts QA pairs and knowledge lists, REWIRE~\citep{nguyen2025recycling} introduces guided rewriting with explicit quality criteria, and later work explores additional target formats such as tutorials, FAQs, and mathematical reformulations~\citep{maini2025beyondweb,niklaus2026finephrase}. Second, on \emph{generator model}, studies spanning models from 270M to 27B parameters suggest that moderate-scale models ($\sim$1B--4B) already produce rephrasings competitive with much larger generators~\citep{maini2024wrap,niklaus2026finephrase}. Third, on \emph{source data quality}, rephrasing can upcycle low-quality web text~\citep{nguyen2025recycling}, although higher-quality source documents still tend to yield stronger downstream performance~\citep{niklaus2026finephrase}. A cross-cutting question concerns how synthetic and original data should be combined, since synthetic-only training often improves factual recall at the expense of broader capabilities, making mixture design important in practice~\citep{maini2024wrap,niklaus2026finephrase}. Despite this progress, existing methods all synthesize from \emph{single documents in isolation}. \ours{} differs from this entire line of work by introducing cross-document knowledge discovery and amplification: instead of rewriting one document at a time, it discovers relational structure from web topology and jointly synthesizes training examples from related entity pairs, explicitly modeling relational knowledge that prior single-document approaches leave untapped.

\section{Conclusion}
\label{sec:conclusion}

We presented \ours{}, a framework that \emph{amplifies} the associative context of factual knowledge by \emph{discovering} cross-document relationships from web topology and synthesizing joint QA over related entity pairs. By mining relational motifs (dual-links and co-mentions) from Wikipedia hyperlinks, \ours{} creates training data with richer relational structure and more diverse retrieval paths than single-document rewriting. On SimpleQA, \ours{} substantially outperforms single-document approaches at 7B and 32B scales, with a favorable scaling trajectory up to 80B tokens.

\bibliography{iclr2025_conference}
\bibliographystyle{iclr2026_conference}

\appendix

\section{Limitation}
Our experiments instantiate \ours{} on Wikipedia, which is a clean and entity-centric corpus. Extending to noisier web corpora (e.g., Common Crawl), where hyperlinks include advertisements, navigation elements, and low-quality references, will require additional filtering heuristics; we are actively exploring this direction.

\section{Training Details}
\label{app:training}

\paragraph{Architecture.}
We use the OLMo-3 architecture~\citep{olmo2025olmo3}, a decoder-only transformer with the Dolma-2 tokenizer (vocabulary size padded to a multiple of 128).
Experiments are conducted at two scales: OLMo-3-7B (7 billion parameters) and OLMo-3-32B (32 billion parameters). We utilize the \texttt{olmo-core} training framework \citep{olmo20242olmo2furious}, which provides a highly optimized and reproducible infrastructure for large-scale distributed training.
All models use FlashAttention-2~\citep{dao2023flashattention2} as the attention backend.

\paragraph{Continued Pretraining.}
All experiments initialize from official OLMo-3 pretrained checkpoints: step 1{,}413{,}814 for 7B and step 679{,}000 for 32B.
We load only the model weights and optimizer state (no trainer state) and continue pretraining on synthetic data mixtures for 1 epoch.
Table~\ref{tab:app_hparams} summarizes the hyperparameters for each configuration.

\begin{table}[h]
\centering
\caption{Continued pretraining hyperparameters. All experiments use a linear decay schedule (no warmup, decaying to 0) and SkipStepAdamW~\citep{olmo2025olmo3} with $\beta_1\!=\!0.9$, $\beta_2\!=\!0.95$, weight decay $0.1$ (embedding weights excluded), max gradient norm $1.0$, and auxiliary z-loss with multiplier $10^{-5}$. Parameters are stored in bfloat16 and gradients are reduced in float32 via HSDP.}
\label{tab:app_hparams}
\scalebox{0.78}{
\begin{tabular}{lccc}
\toprule
\textbf{Hyperparameter} & \textbf{OLMo-3-7B (8B)} & \textbf{OLMo-3-7B (80B)} & \textbf{OLMo-3-32B} \\
\midrule
Sequence length & 8{,}192 & 8{,}192 & 8{,}192 \\
Global batch size (tokens) & $\sim$2M ($2^{21}$) & $\sim$2M ($2^{21}$) & $\sim$4M ($4 \times 2^{20}$) \\
Peak learning rate & $2.07 \times 10^{-5}$ & $2.07 \times 10^{-5}$ & $2.07 \times 10^{-5}$ \\
LR schedule & Linear $\rightarrow$ 0 & Linear $\rightarrow$ 0 & Linear $\rightarrow$ 0 \\
Warmup steps & 0 & 0 & 0 \\
Training steps & 4{,}000 & 40{,}000 & 10{,}000 \\
Training tokens & $\sim$8B & $\sim$80B & $\sim$80B \\
Precision & bfloat16 & bfloat16 & bfloat16 \\
Data parallel & HSDP (block wrap) & HSDP (block wrap) & HSDP (full wrap, shard 64) \\
Activation ckpt. & FFN-only & FFN-only & Budget (50\%) \\
GPUs & 256 $\times$ H20 & 256 $\times$ H20 & 1024 $\times$ H20 \\
\bottomrule
\end{tabular}}
\end{table}

The learning rate of $2.07 \times 10^{-5}$ is inherited from the official OLMo-3 mid-training recipe. We use zero warmup steps because the optimizer state is loaded from the pretrained checkpoint, ensuring stable training from the first step. For the OLMo-3-7B scale, training on $\sim$8B tokens using 256 H20 GPUs takes approximately 4.5 hours with a Model FLOPs Utilization (MFU) of $\sim$65\%. For the larger OLMo-3-32B scale, training on $\sim$80B tokens using 1,024 H20 GPUs requires approximately 1 day and 20 hours. These benchmarks demonstrate the efficiency and scalability of our training pipeline on modern hardware.

\paragraph{Data Format.}
All synthetic QA data is formatted as plain text with ``Question:'' and ``Answer:'' delimiters, consistent with prior WRAP work.
For cross-document QA, the synthesized output directly states facts without referencing source passages, ensuring the model internalizes them as parametric knowledge rather than reading comprehension signals.

\section{Evaluation Details}
\label{app:evaluation_details}

\paragraph{In-Context Learning Setup.} Because our continued pretraining experiments operate on base models (OLMo-3) that have not undergone instruction tuning, these models cannot reliably follow zero-shot formatting directives. To accurately probe their parametric knowledge, we adopt a 5-shot in-context learning protocol for all evaluations, including both SimpleQA and the 12 general benchmarks. Specifically, for each evaluation instance, we prepend the prompt with five demonstration question-answer pairs. For SimpleQA, these demonstrations are sampled directly from the SimpleQA dataset; to ensure strict evaluation integrity and prevent data contamination, any examples used as few-shot demonstrations are explicitly excluded from the active evaluation set during inference.

\paragraph{Decoding and Sampling Parameters.} We adopt distinct decoding strategies tailored to the nature of each benchmark. For \textbf{SimpleQA}, to compute the pass@$k$ metric (where $n=128$), we employ nucleus sampling with a temperature of $0.6$ and top-$p$ of $0.95$ to provide a diverse distribution for knowledge recoverability analysis. In contrast, for the \textbf{12 general benchmarks}, we use greedy decoding (temperature $0.0$) to ensure deterministic and reproducible outputs across all model comparisons. These parameters are held constant across all model scales and data recipes to ensure a fair evaluation.

\paragraph{Unbiased Estimation of pass@$k$.} While the pass@$k$ metric intuitively represents the probability of generating at least one correct answer within $k$ attempts, empirically estimating this by drawing exactly $k$ samples yields high variance. To achieve a more stable and unbiased estimate, we adopt the methodology introduced by \citet{chen2021codex}. For each evaluation instance, we generate $n$ total samples ($n \ge k$) and determine the number of correct responses, $c$. The unbiased estimator for pass@$k$ is then computed as:
\begin{equation}
    \text{pass}@k = 1 - \frac{\binom{n-c}{k}}{\binom{n}{k}}
\end{equation}
where $\binom{\cdot}{\cdot}$ denotes the binomial coefficient. In our experiments, we generate $n=128$ samples per question. This formulation efficiently leverages all $n$ generated samples to calculate the expected pass rate for any evaluation budget $k \le n$, thereby reducing variance without requiring repeated sampling passes.

\section{Synthesis Prompt Templates}
\label{app:prompts}

\paragraph{Single-Document WRAP Prompt (Baseline).}
Following \citet{maini2024wrap}, we use the standard QA-style prompt for single-document synthesis:

\begin{quote}
\small
\texttt{Convert the following paragraph into a conversational format with multiple tags of ``Question:'' followed by ``Answer:''.}
\end{quote}

\paragraph{WRAP++ Cross-Document Joint QA Prompt.}
For cross-document synthesis, we provide the full text of both related entities and apply the following instruction. The prompt enforces three critical constraints: cross-document dependency, explicit factual chaining, and omniscient internalization (see \S\ref{sec:synthesis}).

\begin{quote}
\small
\texttt{You are an expert data generator for language model pretraining.}

\texttt{Below are two related Wikipedia passages:}

\texttt{[Passage A]}\\
\texttt{\{text\_a\}}

\texttt{[Passage B]}\\
\texttt{\{text\_b\}}

\texttt{Task:}\\
\texttt{1) Generate high-quality synthetic QA pairs that REQUIRE information from BOTH Passage A and Passage B to answer.}\\
\texttt{2) The Answer MUST begin with a step-by-step reasoning process. This reasoning must explicitly bridge facts from both passages.}\\
\texttt{3) Do not use external knowledge.}\\
\texttt{4) CRITICAL CONSTRAINT: The generated QA pair will be used to train a model WITHOUT these passages provided as context. Therefore, you MUST act as an omniscient AI stating absolute facts from your own inherent knowledge.}\\
\texttt{\ \ \ - DO NOT use any attribution phrases like `According to Passage A', `Passage B mentions', `As stated in the text', or `Based on the provided documents'.}\\
\texttt{\ \ \ - State the facts directly and confidently.}

\texttt{Output format (strict):}\\
\texttt{Question: [Insert a complex question that bridges facts from both passages]}\\
\texttt{Answer: [Acting as an omniscient AI, directly state all necessary factual premises from both passages, and logically synthesize them to derive the conclusion.]}\\
\texttt{Therefore, [State the final, concise answer.]}
\end{quote}

\paragraph{Synthesis Hyperparameters.}
All QA synthesis uses Qwen3-30B-A3B-Instruct (FP8) as the generator with temperature $0.7$, top-$p$ = $0.8$, and a maximum output length of 32{,}768 tokens.
For the synthesis model scale ablation (\S\ref{sec:ablation_model}), we additionally use Qwen3-235B-A22B-Instruct.

\section{Relation Discovery Statistics}
\label{app:graph_stats}

\begin{table}[h]
\centering
\caption{Statistics of the hyperlink relations discovered from FineWiki (English Wikipedia).}
\label{tab:graph_stats}
\scalebox{0.85}{
\begin{tabular}{lc}
\toprule
\textbf{Statistic} & \textbf{Value} \\
\midrule
Source corpus & FineWiki (English) \\
Raw corpus tokens & $\sim$8.4B \\
Wikipedia articles processed & $\sim$6.7M \\
\midrule
Dual-link pairs ($A \leftrightarrow B$) & $\sim$9.6M \\
Co-mention pairs ($A \rightarrow E \leftarrow B$, $A \rightarrow B$) & $\sim$232M \\
\midrule
Dual-link synthesized tokens & $\sim$3B \\
Co-mention synthesized tokens & $\sim$79.7B \\
Total \ours{} tokens & $\sim$82.7B \\
\midrule
Single-doc WRAP tokens (baseline) & $\sim$5.4B \\
Extended WRAP tokens (baseline) & $\sim$17.4B \\
\bottomrule
\end{tabular}}
\end{table}

Table~\ref{tab:graph_stats} reports the relation discovery and synthesis statistics.
The dual-link motif yields a relatively small but high-precision set of $\sim$9.6M entity pairs, while the co-mention motif provides a much larger pool of $\sim$232M pairs, enabling substantial combinatorial expansion.
The total synthesized corpus of $\sim$82.7B tokens is approximately $15\times$ larger than single-document WRAP on the same source, illustrating the amplification advantage of relation-driven cross-document synthesis.

\section{Mid-Training Evaluation Benchmarks}
\label{app:midtrain_breakdown}

The 12 general tasks used for evaluation in Section~\ref{sec:ablation_midtrain} are categorized as follows:
\begin{itemize}
    \item \textbf{General Knowledge}: ARC \citep{clark2018think}, MMLU Redux \citep{gema2024are}
    \item \textbf{Reading Comprehension}: SQuAD v2 \citep{rajpurkar2018know}, DROP \citep{dua2019drop}
    \item \textbf{Reasoning}: OpenBookQA \citep{mihaylov2018can}, CSQA \citep{talmor2019commonsenseqaquestionansweringchallenge}
    \item \textbf{Natural Language Understanding}: WinoGrande \citep{sakaguchi2021winogrande}, PIQA \citep{bisk2020piqa}, HellaSwag \citep{zellers2019hellaswag}
    \item \textbf{Math}: GSM8K \citep{cobbe2021training}
    \item \textbf{Table Understanding}: WikiTableQuestions \citep{pasupat2015compositional}, TriviaQA \citep{joshi2017triviaqa}
\end{itemize}

Table~\ref{tab:midtrain_perbenchmark} provides per-benchmark results for the mid-training integration experiment described in Section~\ref{sec:ablation_midtrain}.

\begin{table}[h]
\centering
\caption{Per-benchmark breakdown for mid-training integration (OLMo-3-7B, full 100B-token schedule, 3-shot cloze format). Accuracy (\%) is reported for all tasks except SQuAD v2 and DROP, which use token-level F1.}
\label{tab:midtrain_perbenchmark}
\scalebox{0.82}{
\begin{tabular}{llccc}
\toprule
\textbf{Category} & \textbf{Benchmark} & \textbf{Base} & \textbf{Midtrain} & \textbf{\ours{} Mix} \\
\midrule
\multirow{2}{*}{General Knowledge}
 & ARC          & 77.47 & 84.98 & \textbf{85.07} \\
 & MMLU Redux   & 60.30 & 65.30 & \textbf{66.58} \\
\midrule
\multirow{2}{*}{Reading Comprehension}
 & SQuAD v2     & 42.27 & 48.59 & \textbf{49.99} \\
 & DROP         & 40.51 & \textbf{66.51} & 65.36 \\
\midrule
\multirow{2}{*}{Reasoning}
 & OpenBookQA   & 76.40 & \textbf{85.20} & 82.40 \\
 & CSQA         & 72.15 & \textbf{74.20} & 72.97 \\
\midrule
\multirow{3}{*}{Language Understanding}
 & WinoGrande   & 53.28 & \textbf{65.11} & 62.98 \\
 & PIQA         & 73.99 & 74.54 & \textbf{74.92} \\
 & HellaSwag    & 59.83 & 70.15 & \textbf{73.12} \\
\midrule
Math & GSM8K    & 38.36 & \textbf{79.53} & 77.79 \\
\midrule
\multirow{2}{*}{Table \& Trivia}
 & WikiTableQ   & 36.55 & 43.74 & \textbf{44.46} \\
 & TriviaQA     & 62.42 & 61.01 & \textbf{62.23} \\
\midrule
\multicolumn{2}{l}{\textbf{Average}}
 & 57.79 & \textbf{68.24} & 68.16 \\
\bottomrule
\end{tabular}}
\end{table}
\begin{figure*}[h]
    \centering
    \includegraphics[width=\textwidth]{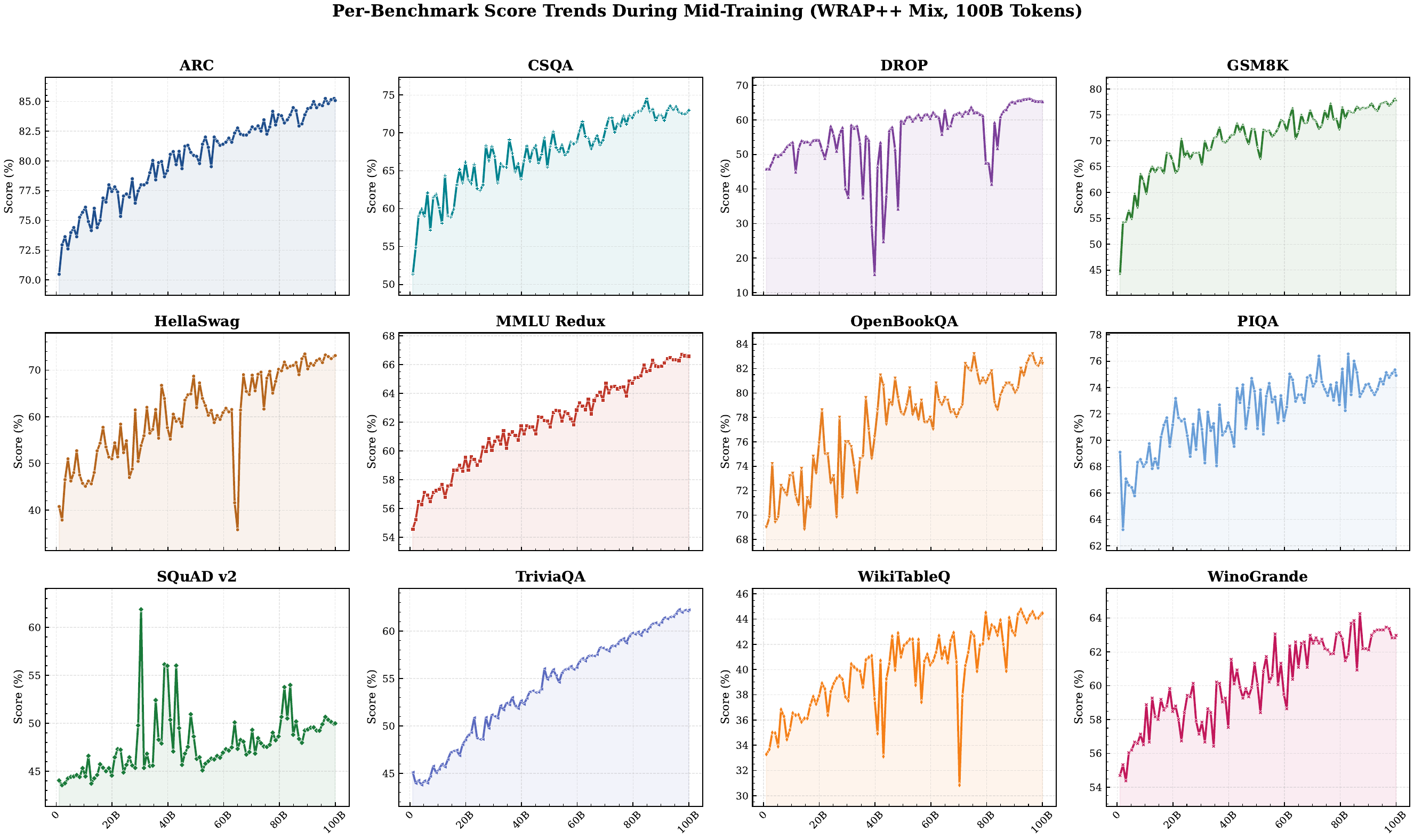}
    \caption{Per-benchmark score trajectories during mid-training with the \ours{} Mix (100B tokens) on OLMo-3-7B. Each subplot tracks one of the 12 evaluation tasks over the course of training. All benchmarks exhibit a clear upward trend, with knowledge-intensive tasks (e.g., HellaSwag, GSM8K, CSQA) showing the most pronounced gains.}
    \label{fig:midtrain_trends}
\end{figure*}

\section{Additional Ablation: Co-Mention with Three Documents}
\label{app:comention_3doc}

In the main co-mention motif ($A \rightarrow E \leftarrow B$ while $A \rightarrow B$), we use the two target entities $A$ and $B$ as input to joint QA synthesis.
A natural extension is to also include the bridging page $E$ as a third input document, potentially providing additional shared context.
We use a dedicated 3-document prompt template that instructs the synthesis model to generate QA requiring facts from all three passages.

\begin{table}[h]
\centering
\caption{Co-mention synthesis using 2 vs.\ 3 input documents (OLMo-3-7B, $\sim$8B tokens, SimpleQA pass@8).}
\label{tab:app_3doc}
\scalebox{0.85}{
\begin{tabular}{lc}
\toprule
\textbf{Co-mention Variant} & \textbf{SimpleQA pass@8} \\
\midrule
2-doc (entities $A$, $B$ only) & \textbf{15.74} \\
3-doc (entities $A$, $B$ + bridge $E$) & 15.42 \\
\bottomrule
\end{tabular}}
\end{table}

Surprisingly, including the bridging page $E$ does not improve---and slightly hurts---performance (Table~\ref{tab:app_3doc}).
We hypothesize that the bridging page introduces distracting context: since $E$ typically links to many entities, its content is broad and may divert the synthesis model from focusing on the specific relationship between $A$ and $B$.
The 2-document formulation used in \ours{} strikes a better balance between relational grounding and synthesis focus.

\section{Synthesized QA Data Statistics}
\label{app:qa_length}

We report comprehensive statistics of the synthesized \ours{} dataset to characterize the length distributions and data composition. Statistics are computed over the full corpus of 240,658,065 QA instances across 24,224 JSONL files, processed in parallel using 256 workers. Character-level and word-level lengths are measured on the raw synthesized text after extracting the ``Question:'' and ``Answer:'' fields from each record.

\paragraph{Dataset Composition.}
Table~\ref{tab:app_data_composition} summarizes the overall dataset composition by relation type.

\begin{table}[h]
\centering
\caption{Composition of the \ours{} synthesized QA dataset by relation type.}
\label{tab:app_data_composition}
\scalebox{0.85}{
\begin{tabular}{lrr}
\toprule
\textbf{Relation Type} & \textbf{QA Instances} & \textbf{Proportion} \\
\midrule
Co-mention & 231,292,954 & 96.1\% \\
Dual-link & 9,365,111 & 3.9\% \\
\midrule
\textbf{Total} & \textbf{240,658,065} & 100.0\% \\
\bottomrule
\end{tabular}}
\end{table}

\begin{figure}[h]
\centering
\includegraphics[width=\linewidth]{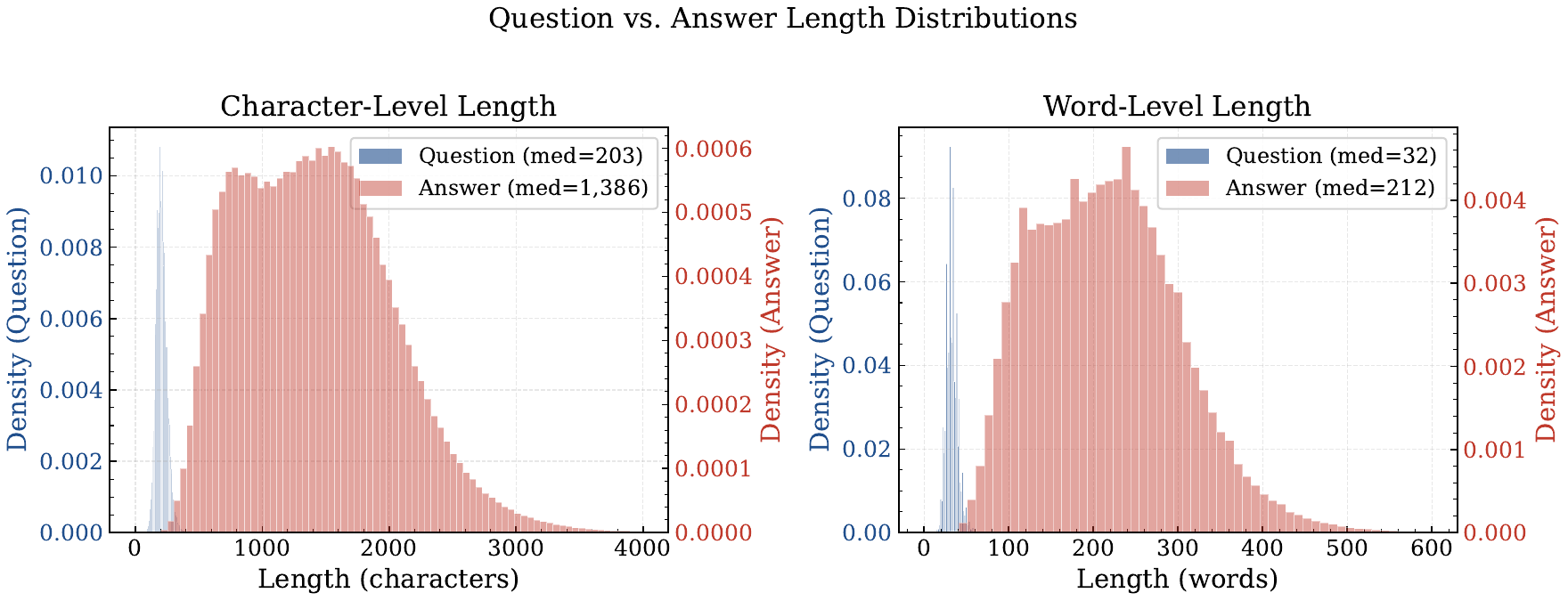}
\caption{Question vs.\ answer length distributions at character level (left) and word level (right). Questions are tightly concentrated around a median of 203 characters (32 words), while answers exhibit a broader, right-skewed distribution with a median of 1,386 characters (212 words), reflecting the explicit factual chaining required by the synthesis prompt.}
\label{fig:q_vs_a_length}
\end{figure}

\paragraph{Length Distributions.}
Table~\ref{tab:app_qa_length_stats} reports the distributional statistics of the synthesized QA text. Questions are concise (median 203 characters, 32 words), while answers are substantially longer (median 1,386 characters, 212 words), reflecting the explicit factual chaining constraint that requires step-by-step reasoning before stating the final conclusion. The overall QA length is concentrated in the 1,000--2,000 character range (56.1\% of all instances), with 27.6\% in the 2,000--5,000 range and 16.0\% in the 500--1,000 range. Fewer than 0.3\% of instances fall outside the 500--5,000 character window, indicating a well-controlled generation process. Figure~\ref{fig:q_vs_a_length} visualizes the question and answer length distributions at both character and word levels.

\begin{table}[h]
\centering
\caption{Length statistics of \ours{} synthesized QA data (aggregated over all 240.7M instances). ``Chars'' denotes character count; ``Words'' denotes whitespace-delimited token count.}
\label{tab:app_qa_length_stats}
\scalebox{0.72}{
\begin{tabular}{l rrrrrrrr}
\toprule
\textbf{Field} & \textbf{Mean} & \textbf{Std} & \textbf{Min} & \textbf{P5} & \textbf{P25} & \textbf{Median} & \textbf{P75} & \textbf{P95} \\
\midrule
\multicolumn{9}{l}{\textit{Character-level}} \\
\quad Question & 207 & 43 & 45 & 143 & 177 & 203 & 233 & 283 \\
\quad Answer & 1{,}424 & 624 & 139 & 563 & 938 & 1{,}386 & 1{,}817 & 2{,}463 \\
\quad QA (combined) & 1{,}651 & 637 & 242 & 766 & 1{,}161 & 1{,}610 & 2{,}051 & 2{,}716 \\
\midrule
\multicolumn{9}{l}{\textit{Word-level}} \\
\quad Question & 33 & 7 & 7 & 23 & 28 & 32 & 37 & 45 \\
\quad Answer & 217 & 91 & 16 & 92 & 149 & 212 & 272 & 364 \\
\bottomrule
\end{tabular}}
\end{table}

\begin{figure}[h]
\centering
\includegraphics[width=0.85\linewidth]{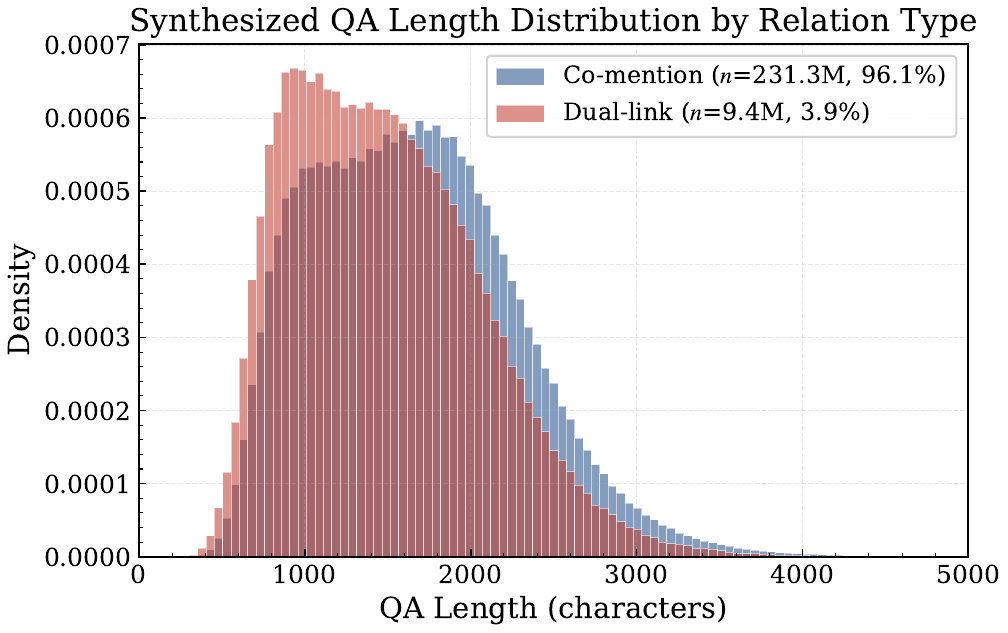}
\caption{Synthesized QA length distribution by relation type. Both subsets exhibit a similar unimodal shape, with the co-mention distribution (blue) shifted slightly rightward relative to dual-link (red), consistent with the longer factual chains required to bridge co-mentioned entities.}
\label{fig:qa_length_by_type}
\end{figure}

\begin{figure}[h]
\centering
\includegraphics[width=\linewidth]{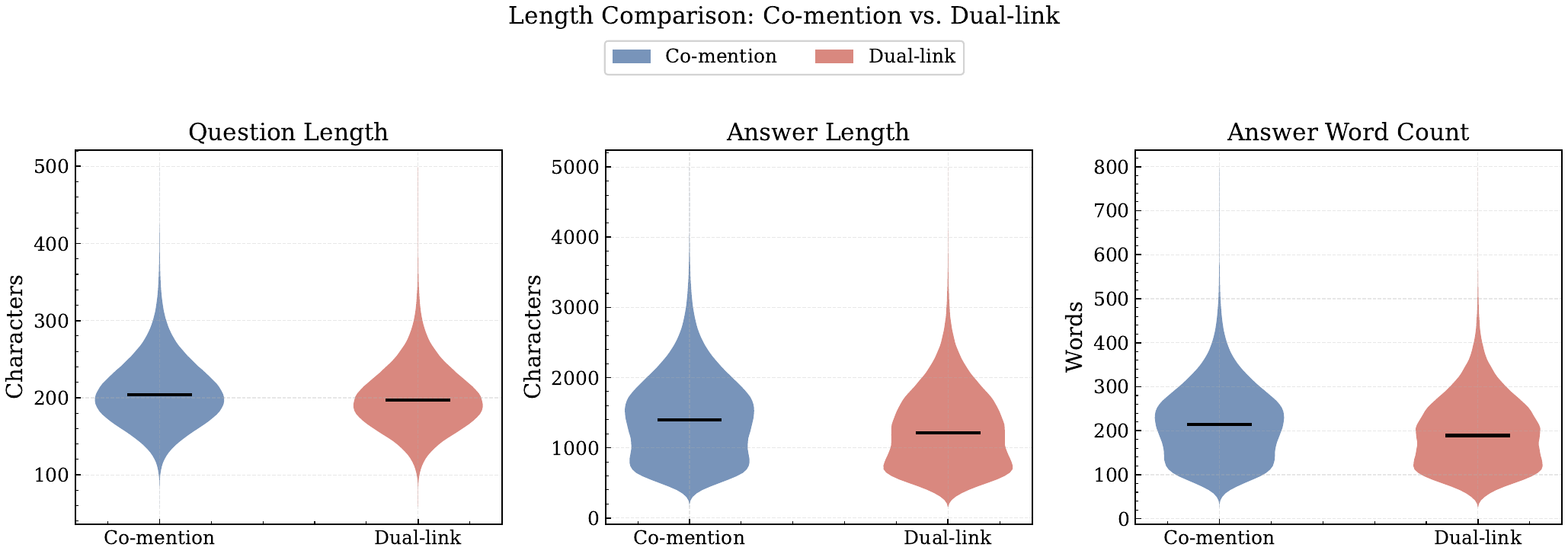}
\caption{Violin plot comparison of question length, answer length, and answer word count between co-mention and dual-link subsets. Black horizontal lines indicate medians. Both motifs produce similarly distributed questions, while co-mention answers are moderately longer, reflecting the additional elaboration needed to bridge indirectly related entities.}
\label{fig:qa_type_comparison}
\end{figure}

\paragraph{Comparison Across Relation Types.}
Table~\ref{tab:app_qa_by_type} compares the QA length characteristics between the two relation motifs. Dual-link instances produce slightly shorter answers (median 1,210 vs.\ 1,394 characters), likely because mutual references tend to encode more focused bilateral relationships, whereas co-mention pairs often involve broader categorical or analogical connections that require more elaboration. Despite these differences, both subsets maintain similar question lengths and overall distributional shape. Figure~\ref{fig:qa_length_by_type} overlays the QA length histograms for both relation types, and Figure~\ref{fig:qa_type_comparison} provides a violin plot comparison across three length dimensions.

\begin{table}[h]
\centering
\caption{Median QA lengths (characters) by relation type. Word-level medians are shown in parentheses.}
\label{tab:app_qa_by_type}
\scalebox{0.85}{
\begin{tabular}{l ccc}
\toprule
\textbf{Relation Type} & \textbf{Question} & \textbf{Answer} & \textbf{QA Total} \\
\midrule
Co-mention & 204 (32 words) & 1{,}394 (213 words) & 1{,}618 \\
Dual-link & 197 (32 words) & 1{,}210 (189 words) & 1{,}428 \\
\midrule
Overall & 203 (32 words) & 1{,}386 (212 words) & 1{,}610 \\
\bottomrule
\end{tabular}}
\end{table}

\begin{figure}[h]
\centering
\includegraphics[width=0.85\linewidth]{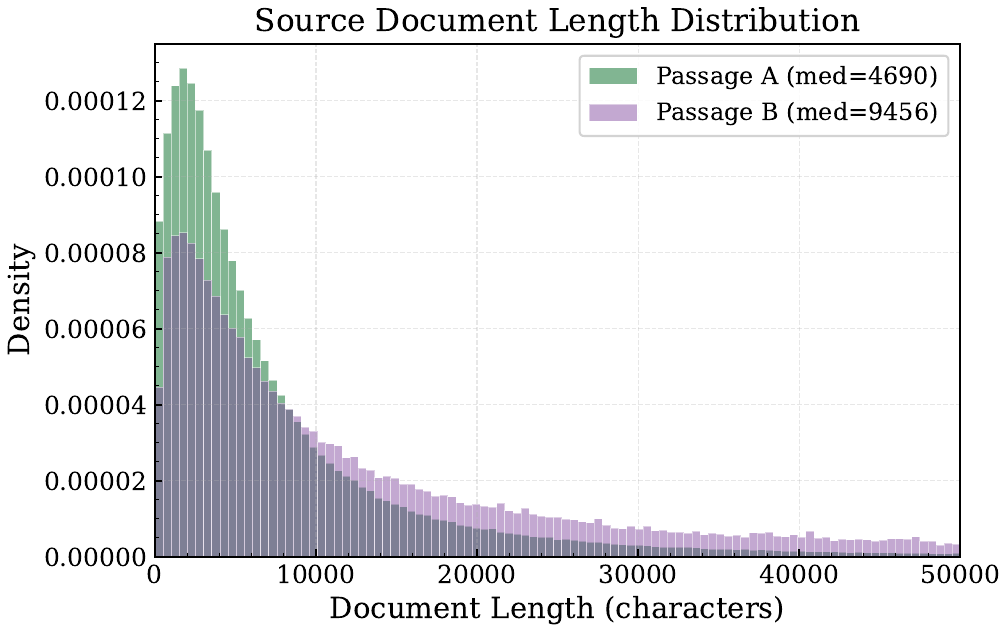}
\caption{Source document length distributions for the two input passages. Passage~A (the referencing entity) is typically shorter (median $\approx$4,700 chars), while Passage~B (the referenced entity) tends to be longer and more prominent (median $\approx$9,500 chars), with a visible mass accumulation at the 50K truncation boundary.}
\label{fig:source_doc_length}
\end{figure}

\paragraph{Source Document Lengths.}
The input Wikipedia passages exhibit substantial length variation. The first passage (\texttt{text\_a}) has a median length of 4,458 characters (P5--P95: 519--27,963), while the second passage (\texttt{text\_b}) is generally longer with a median of 9,575 characters (P5--P95: 943--50,004, where 50,004 indicates truncation at the maximum context window). This asymmetry arises because co-mention pairs order documents by the directed edge $A \rightarrow B$, where $B$ (the referenced entity) tends to be a more prominent article. Figure~\ref{fig:source_doc_length} visualizes this distributional asymmetry.

\begin{figure}[h]
\centering
\includegraphics[width=\linewidth]{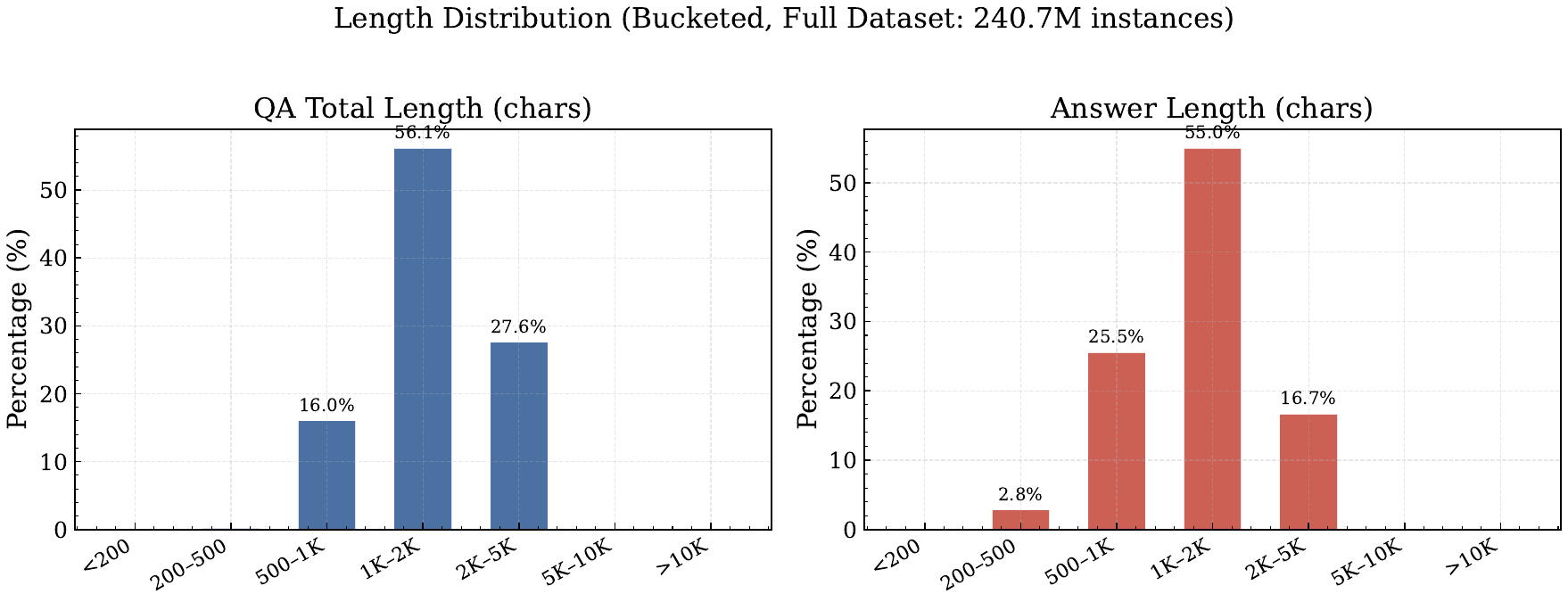}
\caption{Bucketed length distributions computed over the full dataset of 240.7M instances. Left: combined QA length; right: answer-only length. The 1K--2K character bucket dominates both distributions ($>$55\%), with a secondary concentration in the 2K--5K range for QA and the 500--1K range for answers. Extreme lengths ($>$5K or $<$200) are negligible.}
\label{fig:length_bucketed_bar}
\end{figure}

\paragraph{QA Length Bucketed Distribution.}
Table~\ref{tab:app_qa_buckets} and Figure~\ref{fig:length_bucketed_bar} provide a bucketed view of the combined QA and answer length distributions. The synthesis process produces a unimodal distribution with the majority of instances in the 1,000--2,000 character range. No instances have empty answers, and fewer than 0.01\% of answers are shorter than 200 characters, confirming that the explicit factual chaining constraint effectively prevents degenerate outputs.

\begin{table}[h]
\centering
\caption{Bucketed length distributions for the combined QA text and answer text (character-level).}
\label{tab:app_qa_buckets}
\scalebox{0.78}{
\begin{tabular}{l rr rr}
\toprule
\multirow{2}{*}{\textbf{Length Range (chars)}} & \multicolumn{2}{c}{\textbf{QA (combined)}} & \multicolumn{2}{c}{\textbf{Answer only}} \\
\cmidrule(lr){2-3} \cmidrule(lr){4-5}
& \textbf{Count} & \textbf{\%} & \textbf{Count} & \textbf{\%} \\
\midrule
$[0, 200)$       & 0 & 0.0 & 2{,}479 & ${<}0.01$ \\
$[200, 500)$     & 456{,}399 & 0.2 & 6{,}771{,}177 & 2.8 \\
$[500, 1{,}000)$ & 38{,}609{,}776 & 16.0 & 61{,}411{,}377 & 25.5 \\
$[1{,}000, 2{,}000)$ & 135{,}127{,}249 & 56.1 & 132{,}242{,}235 & 55.0 \\
$[2{,}000, 5{,}000)$ & 66{,}336{,}557 & 27.6 & 40{,}114{,}788 & 16.7 \\
$[5{,}000, 10{,}000)$ & 99{,}207 & ${<}0.1$ & 89{,}225 & ${<}0.1$ \\
$[10{,}000, \infty)$ & 28{,}877 & ${<}0.1$ & 26{,}784 & ${<}0.1$ \\
\bottomrule
\end{tabular}}
\end{table}

\section{Qualitative Analysis}
\label{app:qualitative}

\paragraph{Example: WRAP vs.\ WRAP++ Synthesis.}

\textbf{Source entity A:} Ludwig G\"{o}ransson (Swedish composer, film score artist)\\
\textbf{Source entity B:} Oppenheimer (2023 film directed by Christopher Nolan)

\begin{table*}[t]
\centering
\small
\caption{Qualitative comparison between single-document WRAP and cross-document \ours{} synthesis for the Ludwig G\"{o}ransson and \emph{Oppenheimer} example.}
\label{tab:qualitative_compare}
\begin{tabular}{@{}p{0.47\textwidth}@{\hspace{0.03\textwidth}}p{0.47\textwidth}@{}}
\toprule
\textbf{WRAP (single-document QA on entity A)} & \textbf{\ours{} (cross-document QA on entities A+B)} \\
\midrule
\begin{minipage}[t]{\linewidth}
\raggedright
\textbf{Q:} Who is Ludwig G\"{o}ransson?\par
\textbf{A:} Ludwig G\"{o}ransson is a Swedish composer known for film scores.\par\vspace{0.5em}
\textbf{Q:} What films has G\"{o}ransson scored?\par
\textbf{A:} He has scored \emph{Black Panther}, \emph{Tenet}, and \emph{Oppenheimer}.
\end{minipage}
&
\begin{minipage}[t]{\linewidth}
\raggedright
\textbf{Q:} What films earned Nolan's original Tenet composer and his replacement their second Oscars?
\textbf{A:} Hans Zimmer turned down Tenet for Dune, which earned his second Oscar (after The Lion King). His replacement, Ludwig Göransson, later won his second Oscar for Oppenheimer (after Black Panther). Therefore, the answer is Dune and Oppenheimer.
\end{minipage}
\\
\bottomrule
\end{tabular}
\end{table*}

As shown in Table~\ref{tab:qualitative_compare}, the \ours{} output creates richer associative context: it contrasts with a commonly confused entity (Zimmer), provides cross-film comparisons, and generates reverse-direction queries---all contributing to more robust knowledge encoding.

\end{document}